\documentclass[10pt]{article}

\usepackage{graphicx}
\usepackage{amsmath}
\usepackage{changepage}
\usepackage{authblk}
\usepackage{listings}

\usepackage{xpatch}
\xpatchcmd{\author}{\relax#1\relax}{\relax\detokenize{#1}\relax}{}{}

\begin{document}
\title{Randomized Iterative Reconstruction for Sparse View X-ray Computed Tomography}
\author[\empty]{D. Trinca\textsuperscript{a,}\thanks{Corresponding author, email: dntrinca@yahoo.com}}
\author[b]{Y. Zhong}
\affil[a]{Sc Piretus Prod Srl, Osoi, Romania}
\affil[b]{Russian-Chinese Laboratory of Radiation Control and Inspection, Tomsk Polytechnic University, Russian Federation}
\date{}
\maketitle
\begin{abstract}
With the availability of more powerful computers, iterative reconstruction algorithms are the subject of an ongoing work in the design of more efficient reconstruction algorithms for X-ray computed tomography. In this work, we show how two analytical reconstruction algorithms can be improved by correcting the corresponding reconstructions using a randomized iterative reconstruction algorithm. The combined analytical reconstruction followed by randomized iterative reconstruction can also be viewed as a reconstruction algorithm which, in the experiments we have conducted, uses up to $35\%$ less projection angles as compared to the analytical reconstruction algorithms and produces the same results in terms of quality of reconstruction, without increasing the execution time significantly.
\end{abstract}
\section{Introduction}
Iterative reconstruction algorithms \cite{1,4,5,6,7} for X-ray computed tomography (CT) \cite{1,2} have recently been considered in the designing of more efficient reconstruction methods in X-ray CT, especially for the task of obtaining high quality reconstructed tomograms when using a sparse view scanning experiment \cite{4,5,6,7}.

Comparing to the conventional filtered back-projection algorithm \cite{2,3}, which doesn't obtain good results for sparse view scanning, iterative reconstruction algorithms normally take significantly more time for obtaining reconstructed tomograms of comparable accuracy. Most work for sparse view X-ray computed tomography has considered the use of Total Variation (TV) optimization for obtaining reconstruction of good accuracy, but the problem is that in general they report long execution times \cite{4,5,6,7}, usually in the interval 20 - 60 seconds. In this work, we show how two analytical reconstruction algorithms (the filtered back-projection algorithm and the recently proposed direct integration of the inverse Radon transform \cite{3}) can be improved by correcting the corresponding reconstructions using a randomized iterative reconstruction algorithm. The combined analytical reconstruction followed by randomized iterative reconstruction can also be viewed as a reconstruction algorithm which, in the experiments we have conducted, uses up to $35\%$ less projection angles as compared to the analytical reconstruction algorithms and produces the same results in terms of quality of reconstruction, without increasing the execution time significantly.

The work is organized as follows. First, we describe an example that we use and show the corresponding results obtained by the filtered back-projection algorithm and the direct integration method. Then, we present the details of the proposed randomized iterative reconstruction algorithm, along with the produced reconstructions. At the end, we discuss possible improvements in terms of execution time and reconstruction accuracy.
\section{Filtered Back-Projection}
\begin{figure}[t]
\centering
\includegraphics[width=250pt]{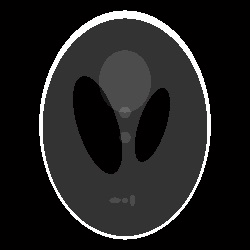}
\caption{Cross-section image of size 250 by 250 pixels}
\label{fig:1}
\end{figure}

\begin{figure}[t]
\centering
\includegraphics[width=250pt]{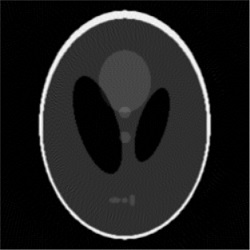}
\caption{Result of reconstruction by the filtered back-projection algorithm, 270 projection angles}
\label{fig:fbp270}
\end{figure}

The iterative reconstruction algorithms for X-ray CT are typically compared to the filtered back-projection algorithm. As an  example that we will use for comparison, consider the Shepp-Logan cross-section shown in Fig. \ref{fig:1}, of size 250 by 250 pixels.
For this cross-section, consider the following parameters:
\begin{enumerate}
\item distance from fan-beam source to origin of rotation of inspected object = 800  mm,
\item distance from fan-beam source to line of detectors = 1500 mm,
\item number of detectors equally spaced (with 1mm for each detector) on the detector line = 359,
\item number of projection angles (views) = 270.
\end{enumerate}
The corresponding sinogram is a matrix with 359 lines and 270 columns. For this set of parameters, the result obtained by the MATLAB implementation of the filtered back-projection algorithm is shown in Fig. \ref{fig:fbp270}.
\section{Direct Integration of the Inverse Radon Transform}
\begin{figure}[t]
\centering
\includegraphics[width=250pt]{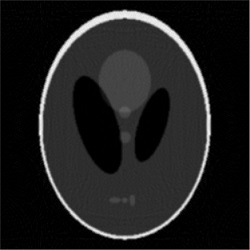}
\caption{Result of reconstruction by the direct integration method, 270 projection angles}
\label{fig:cda270}
\end{figure}

The filtered back-projection algorithm is based on the Fourier Slice Theorem, which makes use of Fourier transform. However, very recently, we have proposed a method of directly (that is, without Fourier transform) calculating the inverse Radon transform \cite{3}. The general equation of the inverse Radon transform is of the form
\begin{equation}
\mu(r,x,y)=-\frac{1}{\pi^{2}}\int\limits_{0}^{2\pi}d\varphi\int\limits_{-L}^{L}\frac{\partial{S}(p,\varphi)}{\partial p}\frac{dp}{q(p,\varphi,x,y)},
\label{eq:iRt1}
\end{equation}
where $q(p,\varphi,x,y)$ has to be defined for each scanning geometry (i.e., parallel rays, fan-beam, etc.) separately. For the fan-beam case that we consider in this paper, the inverse Radon transform has form
\begin{equation}
\mu(x,y)=\frac{1}{\pi^{2}}\int\limits_{0}^{2\pi}\frac{Rd\varphi}{Y}\int\limits_{-1}^{1}\sqrt{1+\frac{p^{2}}{R^{2}}}\frac{\partial{S}(p,\varphi)}{\partial p}\frac{dp}{p-\frac{RX}{Y}}.
\label{eq:iRt2}
\end{equation}
We have presented in \cite{3} a method of calculating (by direct approximation) the right-hand side of (\ref{eq:iRt2}), and showed that it obtains more accurate reconstruction as compared to the filtered back-projection algorithm (for the case of sparse view scanning with at most 180 projection angles). The result obtained through this direct integration method, for the example with 270 projection angles, is shown in Fig. \ref{fig:cda270}.
\section{Randomized Iterative Reconstruction}
In this section, we propose two simple but, as we shall see, effective randomized iterative reconstruction algorithms for X-ray CT. Both provide about the same reconstruction accuracy, although we shall see that the first one might possibly fail to work, while the second one works successfully and obtains very good results.

The first algorithm works as follows. Let $\mu$ be the density function to be reconstructed from sinogram $S$, ${\it nd}$ the number of detectors, and ${\it np}$ the number of projection angles (views). The matrix $\mu$ is initialized, before iterations start, with the solution obtained by other algorithm (say the filtered back-projection algorithm or the direct integration method). At each iteration, we choose a random detector ${\it d}$ and a random projection angle ${\it p}$, $1\leq{\it d}\leq{\it nd}$ and $1\leq{\it p}\leq{\it np}$. For these random detector and projection angle, consider the X-ray that goes from source to detector number ${\it d}$, for the ${\it p}$-th projection angle. Consider the elements of the current reconstruction $\mu$ that correspond to this randomly chosen X-ray. To update these elements, let ${\it li}[{\it d},{\it p}]$ be the line integral of these elements along the current X-ray. If
\begin{displaymath}
\frac{{\it li}[{\it d},{\it p}]}{S[{\it d},{\it p}]}\neq{1.0},
\end{displaymath}
which means that the line integral of the chosen X-ray doesn't equal the corresponding value in the sinogram, then we update the elements in $\mu$ corresponding to the current X-ray by multiplying each of them with the ratio
\begin{displaymath}
\frac{S[{\it d},{\it p}]}{{\it li}[{\it d},{\it p}]}.
\end{displaymath}

The second algorithm works as follows. The matrix $\mu$ is initialized, before iterations start, with the solution obtained by other algorithm (say filtered back-projection or the direct integration method). At the current iteration, we choose randomly two X-rays, the first being determined by the randomly chosen detector ${\it d1}$ and projection angle ${\it p1}$, and the second by the randomly chosen detector ${\it d2}$ and projection angle ${\it p2}$, such that the two chosen X-rays don't have any element in common. For the chosen X-rays, we consider two distinct cases: (A) both sinogram values $S[{\it d1},{\it p1}]$ and $S[{\it d2},{\it p2}]$ are strictly positive or (B) at least one of them is equal to 0.0. For the case (A), we check if
\begin{displaymath}
\frac{{\it li}[{\it d1},{\it p1}]}{{\it li}[{\it d2},{\it p2}]}=\frac{S[{\it d1},{\it p1}]}{S[{\it d2},{\it p2}]},
\end{displaymath}
i.e., we check to see if the ratio of the corresponding sinogram values is equal to the ratio of the line integrals of the chosen X-rays. If yes, it means that no update is needed. Otherwise, if the two ratios are not equal, we update the elements in $\mu$ corresponding to the two X-rays chosen. We propose to update by first calculating the number $x$ such that
\begin{equation}
\frac{{\it li}[{\it d1},{\it p1}]+x}{{\it li}[{\it d2},{\it p2}]-x}=\frac{S[{\it d1},{\it p1}]}{S[{\it d2},{\it p2}]},
\label{eq:x}
\end{equation}
which means that
\begin{displaymath}
x=\frac{\dfrac{S[{\it d1},{\it p1}]}{S[{\it d2},{\it p2}]}*{\it li}[{\it d2},{\it p2}]-{\it li}[{\it d1},{\it p1}]}{1.0+\dfrac{S[{\it d1},{\it p1}]}{S[{\it d2},{\it p2}]}}.
\end{displaymath}
This update means that each element $e$ in $\mu$ corresponding to the first X-ray is replaced with
\begin{displaymath}
e+x*\frac{e}{{\it li}[{\it d1},{\it p1}]},
\end{displaymath}
and  each element $e$ in $\mu$ corresponding to the second X-ray is replaced with
\begin{displaymath}
e-x*\frac{e}{{\it li}[{\it d2},{\it p2}]}.
\end{displaymath}
For example, if the elements corresponding to the first X-ray are $e_{1},e_{2},e_{3}$ and ${\it seg}_{1},{\it seg}_{2},{\it seg}_{3}$ are the corresponding segments and
\begin{displaymath}
{\it li}[{\it d1},{\it p1}] = {\it seg}_{1}*e_{1}+{\it seg}_{2}*e_{2}+{\it seg}_{3}*e_{3} = \sum_{i=1}^{3}{\it seg}_{i}*e_{i},
\end{displaymath}
then 
\begin{align}
 \sum_{i=1}^{3}{\it seg}_{i}*(e_{i}+x*\frac{e_{i}}{{\it li}[{\it d1},{\it p1}]}) &=  \sum_{i=1}^{3}{\it seg}_{i}*e_{i}+ \sum_{i=1}^{3}{\it seg}_{i}*x*\frac{e_{i}}{{\it li}[{\it d1},{\it p1}]}\nonumber\\
&= \sum_{i=1}^{3}{\it seg}_{i}*e_{i}+ x*\frac{\sum_{i=1}^{3}{\it seg}_{i}*e_{i}}{{\it li}[{\it d1},{\it p1}]}\nonumber\\
&= {\it li}[{\it d1},{\it p1}]+x*\frac{{\it li}[{\it d1},{\it p1}]}{{\it li}[{\it d1},{\it p1}]}\nonumber\\
&= {\it li}[{\it d1},{\it p1}]+x,\nonumber
\end{align}
which is exactly the numerator of the first fraction in relation (\ref{eq:x}). For the second X-ray, if the elements are $e_{4},e_{5},e_{6}$ and ${\it seg}_{4},{\it seg}_{5},{\it seg}_{6}$ are the corresponding segments and
\begin{displaymath}
{\it li}[{\it d2},{\it p2}] = {\it seg}_{4}*e_{4}+{\it seg}_{5}*e_{5}+{\it seg}_{6}*e_{6} = \sum_{i=4}^{6}{\it seg}_{i}*e_{i},
\end{displaymath}
then 
\begin{align}
 \sum_{i=4}^{6}{\it seg}_{i}*(e_{i}-x*\frac{e_{i}}{{\it li}[{\it d2},{\it p2}]}) &=  \sum_{i=4}^{6}{\it seg}_{i}*e_{i}-\sum_{i=4}^{6}{\it seg}_{i}*x*\frac{e_{i}}{{\it li}[{\it d2},{\it p2}]}\nonumber\\
&= \sum_{i=4}^{6}{\it seg}_{i}*e_{i}-x*\frac{\sum_{i=4}^{6}{\it seg}_{i}*e_{i}}{{\it li}[{\it d2},{\it p2}]}\nonumber\\
&= {\it li}[{\it d2},{\it p2}]-x*\frac{{\it li}[{\it d2},{\it p2}]}{{\it li}[{\it d2},{\it p2}]}\nonumber\\
&= {\it li}[{\it d2},{\it p2}]-x,\nonumber
\end{align}
which is exactly the denominator of the first fraction in relation (\ref{eq:x}). For the case (B), we update the corresponding elements of each X-ray that has sinogram value of 0.0, by setting them to 0.0.

Each of the algorithms stops when the current situation in the reconstruction $\mu$ is in very close agreement with the sinogram. For our example, we have chosen to stop the number of iterations by visual inspection of the reconstruction, but precise stopping criteria have to be defined in practice.

\underline{\it the first algorithm vs the second algorithm:} The first algorithm presents an important issue, which can be explained as follows. The initial solution, before iterations start, needs to have the property that its values are in close agreement with the attenuation coefficients of the practical experiment when the sinogram has been obtained. This has to be the case, because at each iteration the current situation (line integral) of the chosen X-ray has to be in close agreement with the corresponding value in the sinogram. Otherwise stated, the initial solution needs to have the property that at each iteration the ratio
\begin{displaymath}
\frac{{\it li}[{\it d},{\it p}]}{S[{\it d},{\it p}]}
\end{displaymath}
is close enough to $1.0$, because otherwise the updating from an iteration to the next one could create very small or very large numbers in $\mu$ (beyond their possible range). Scaling an initial solution such that each possible line integral is in close agreement with the corresponding value in the sinogram adds additional reconstruction time needed. The good thing with the second algorithm is that such an initial scaling before iterations start is eliminated, as at each iteration of we don't compare the line integral of a chosen X-ray with the corresponding value in the sinogram. Some initial solutions normally return result very close to the actual attenuation coefficients, but others don't, which means that the second algorithm is preferable over the first one, as such initial scaling is not needed.

\underline{\it common elements:} Regarding the second algorithm, we have stated that when the two X-rays are chosen, they should not have any element of the reconstruction in common. That is a very important condition that needs to be satisfied, otherwise the formulas presented don't work. The reason is the following: if they would have any element in common, say the element is $e$, then after updating the element $e$ when dealing with the first X-ray, a second update on it would occur, when dealing with the second X-ray.

\underline{\it execution time:} For the second algorithm, the execution time can be significantly improved as follows. First, for the specific geometry and parameters of the experiment, precomputed pairs ((detector1,projection1),(detector2,projection2))  can be just loaded into memory from a database, such that during the algorithm there is no need to generate such pairs. Second, when dealing with the cases (A) and (B) we can do as follows. Before the algorithm starts, we can build a set ${\it set}_{z}$ such that an element $e$ of $\mu$ is in ${\it set}_{z}$ if: (a) there is at least one detector-projection pair (d,p) such that the X-ray corresponding to (d,p) goes through $e$, and (b) the sinogram value $S[{\it d},{\it p}]$ is 0.0. Then, when having the set ${\it set}_{z}$, the algorithm is executed as described, but with the following two modifications: (modification 1) the line integrals at each iteration are calculated by considering only the elements that are not in ${\it set}_{z}$, because these elements should remain 0.0; (modification 2) whenever case (B) occurs at the current iteration, nothing is done, which means that the current iteration is not counted. With this improvement, the execution time of the second randomized algorithm, for the parameters that we have set and reconstructed image of size 250 by 250 pixels, is about 0.1 seconds on a good multi-core computer (Intel Core i5-7600K @ 3.80 GHz).
\section{Results of Correction}
\begin{figure}[t]
\centerline{
\includegraphics[width=505pt]{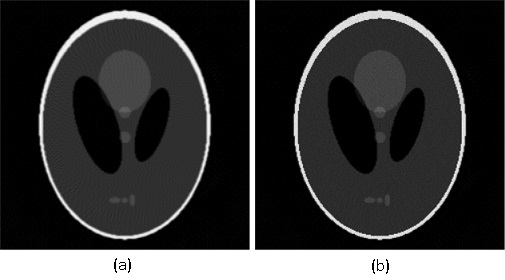}
}
\caption{(a) result of filtered back-projection (i.e., initial solution) (b) the correction obtained by running the second randomized algorithm with 125000 iterations}
\label{fig:fbp125000}
\end{figure}

\begin{figure}[t]
\centerline{
\includegraphics[width=505pt]{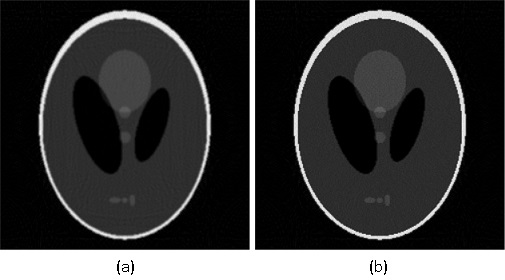}
}
\caption{(a) result of the direct integration method (i.e., initial solution) (b) the correction obtained by running the second randomized algorithm with 125000 iterations}
\label{fig:cda125000}
\end{figure}

\begin{figure}[t]
\centerline{\includegraphics[width=250pt]{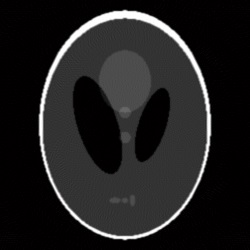}}
\caption{result of filtered back-projection when using 360 projection views instead of 270}
\label{fig:fbp360}
\end{figure}

\begin{figure}[t]
\centerline{\includegraphics[width=250pt]{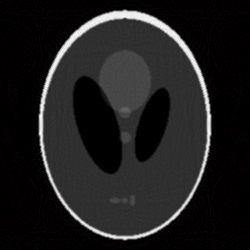}}
\caption{result of the direct integration method when using 360 projection views instead of 270}
\label{fig:cda360}
\end{figure}

Given the problems we have pointed out regarding the first randomized algorithm, we have focused on the results obtained by the second randomized algorithm. For the parameters that we have fixed above, the results obtained by running the second randomized algorithm with 125000 iterations are given in Fig. \ref{fig:fbp125000} and Fig. \ref{fig:cda125000} (Fig. \ref{fig:fbp125000} (a) is Fig. \ref{fig:fbp270}, and Fig. \ref{fig:cda125000} (a) is Fig. \ref{fig:cda270}). In both cases, the correction is significant, and the result of correction is very close in terms of reconstruction quality to the result obtained by running each of the two analytical algorithms alone with full view of 360 projection angles. That is, the result shown in Fig. \ref{fig:fbp125000} (b) is comparable to the result shown in Fig. \ref{fig:fbp360}, and the result shown in Fig. \ref{fig:cda125000} (b) is comparable to the result shown in Fig. \ref{fig:cda360}, which means a reduction in the number of views of $25\%$ (in Figs. \ref{fig:fbp360} and \ref{fig:cda360} the reconstruction is more homogeneous, but with more visible artifacts). From the experiments we have conducted, we have remarked that the number of views can be reduced up to about $35\%$ and the results are still comparable to the results obtained by the analytical algorithms with full view, provided an appropriate number of iterations is chosen.
\section{Visual C++ Implementation}
\subsection{Randomized Iterative Reconstruction}
We have used Visual Studio Enterprise 2015 with Updates, 64 bits, installed on a Windows 8.1 64 bits operating system, for the implementation and testing of the second proposed randomized algorithm. We provide below the main file of the implementation. The project has been created as follows:
\begin{enumerate}
\item In Visual Studio, a new project is created, with the name "im1(VS2015-Enterprise-with-Updates)" by using the 'New' option from the 'File' menu, then choosing 'Project..', and then choosing 'Win32 Project';
\item After the project is created, we have created a new menu item called 'X-ray Testing', with the option 'Iterative Method 1'; The ID of this menu option is thus 'ID\_X\_ITERATIVEMETHOD1';
\item As one can see in the code, in the function WndProc, this menu option is treated accordingly and the whole reconstruction procedure happens inside it.
\end{enumerate}
So, basically, there are only 2 main changes to the default project files created in Visual Studio: first, the menu option 'Iterative Method 1' is created; second, the main cpp file of the project is modified according to the code shown. The results that we obtained are for the Release version of the project, with 64 bits compilation.
\lstset {language=C++}
\begin{lstlisting}
// im1(VS2015-Enterprise-with-Updates).cpp
//

#include "stdafx.h"
#include "im1(VS2015-Enterprise-with-Updates,v1).h"
#define _USE_MATH_DEFINES
#include <cmath>
#include <time.h>

#include <vector>
#include <set>
#include <algorithm>

using namespace std;

#define MAX_LOADSTRING 100

// Global Variables:
HINSTANCE hInst;
WCHAR szTitle[MAX_LOADSTRING];
WCHAR szWindowClass[MAX_LOADSTRING];

const unsigned int np   = 359;
const unsigned int nf   = 270;
const unsigned int nx   = 250;
const unsigned int ny   = 250;
const double       a    = 700;
const double       d    = 800;
const double       c1   = (2.0*M_PI) / nf;
const unsigned int npnf = np * nf;
const unsigned int nxny = nx * ny;

struct INF {
   unsigned int *ind;
   unsigned int *Ind;
   double *seg;
   double *Seg;
   double SumOfSegs;
   unsigned int  count;
   unsigned int  Count;
};

HDC              hdcMem;
HGDIOBJ          hbmOld;
BITMAPINFOHEADER bmih;
BITMAPINFO       dbmi;
HBITMAP          hbmp = NULL;
BITMAP           bmp;

double duration1 = 0.0;
double duration2 = 0.0;
double min1      = 0.0;
double max1      = 0.0;

unsigned char *pixels = NULL;
void          *bits   = NULL;

unsigned int   Status = 0;

// Forward declarations of functions included in this code module:
ATOM                MyRegisterClass(HINSTANCE hInstance);
BOOL                InitInstance(HINSTANCE, int);
LRESULT CALLBACK    WndProc(HWND, UINT, WPARAM, LPARAM);
INT_PTR CALLBACK    About(HWND, UINT, WPARAM, LPARAM);

int APIENTRY wWinMain(_In_     HINSTANCE hInstance,
	              _In_opt_ HINSTANCE hPrevInstance,
	              _In_        LPWSTR lpCmdLine,
	              _In_           int nCmdShow)
{
  UNREFERENCED_PARAMETER(hPrevInstance);
  UNREFERENCED_PARAMETER(lpCmdLine);

  // Initialize global strings
  LoadStringW(hInstance, IDS_APP_TITLE, szTitle, MAX_LOADSTRING);
  LoadStringW(hInstance, IDC_IM1VS2015ENTERPRISEWITHUPDATESV1, szWindowClass, MAX_LOADSTRING);
  MyRegisterClass(hInstance);

  // Perform application initialization:
  if (!InitInstance(hInstance, nCmdShow))
  {
    return FALSE;
  }

  HACCEL hAccelTable = LoadAccelerators(hInstance,
                                          MAKEINTRESOURCE(IDC_IM1VS2015ENTERPRISEWITHUPDATESV1));

  MSG msg;

  // Main message loop:
  while (GetMessage(&msg, NULL, 0, 0))
  {
    if (!TranslateAccelerator(msg.hwnd, hAccelTable, &msg))
    {
      TranslateMessage(&msg);
      DispatchMessage(&msg);
    }
  }

  return (int)msg.wParam;
}



//
//  FUNCTION: MyRegisterClass()
//
//  PURPOSE: Registers the window class.
//
ATOM MyRegisterClass(HINSTANCE hInstance)
{
  WNDCLASSEXW wcex;

  wcex.cbSize = sizeof(WNDCLASSEX);

  wcex.style = CS_HREDRAW | CS_VREDRAW;
  wcex.lpfnWndProc = WndProc;
  wcex.cbClsExtra = 0;
  wcex.cbWndExtra = 0;
  wcex.hInstance = hInstance;
  wcex.hIcon = LoadIcon(hInstance, MAKEINTRESOURCE(IDI_IM1VS2015ENTERPRISEWITHUPDATESV1));
  wcex.hCursor = LoadCursor(NULL, IDC_ARROW);
  wcex.hbrBackground = (HBRUSH)(COLOR_WINDOW + 1);
  wcex.lpszMenuName = MAKEINTRESOURCEW(IDC_IM1VS2015ENTERPRISEWITHUPDATESV1);
  wcex.lpszClassName = szWindowClass;
  wcex.hIconSm = LoadIcon(wcex.hInstance, MAKEINTRESOURCE(IDI_SMALL));

  return RegisterClassExW(&wcex);
}

//
//   FUNCTION: InitInstance(HINSTANCE, int)
//
//   PURPOSE: Saves instance handle and creates main window
//
//   COMMENTS:
//
//        In this function, we save the instance handle in a global variable and
//        create and display the main program window.
//
BOOL InitInstance(HINSTANCE hInstance, int nCmdShow)
{
  hInst = hInstance; // Store instance handle in our global variable

  HWND hWnd = CreateWindowW(szWindowClass, szTitle, WS_OVERLAPPEDWINDOW, CW_USEDEFAULT, 0,
                                                  CW_USEDEFAULT, 0, NULL, NULL, hInstance, NULL);

  if (!hWnd)
  {
    return FALSE;
  }

  ShowWindow(hWnd, SW_SHOWMAXIMIZED);
  UpdateWindow(hWnd);

  return TRUE;
}

//
//  FUNCTION: WndProc(HWND, UINT, WPARAM, LPARAM)
//
//  PURPOSE:  Processes messages for the main window.
//
//  WM_COMMAND  - process the application menu
//  WM_PAINT    - Paint the main window
//  WM_DESTROY  - post a quit message and return
//
//
LRESULT CALLBACK WndProc(HWND hWnd, UINT message, WPARAM wParam, LPARAM lParam)
{
  switch (message)
  {
    case WM_COMMAND:
    {
      int wmId = LOWORD(wParam);
      // Parse the menu selections:
      switch (wmId)
      {
        case ID_X_ITERATIVEMETHOD1:
        {
          SetCursor(LoadCursor(NULL, IDC_WAIT));
          InvalidateRect(hWnd, NULL, TRUE);
          Status = 0;
          MSG msg;
          msg.hwnd = hWnd;
          msg.message = WM_PAINT;
          DispatchMessage(&msg);

          clock_t start1;
          clock_t start2;
          clock_t finish;

          start1 = clock();

          unsigned int i;
          unsigned int j;
          unsigned int k;
          INF         *Z1 = NULL;
          double      *Mu = NULL;
          double      *RI = NULL;
          double      *O1 = NULL;


          Z1 = (INF*   )malloc(npnf * sizeof(INF   ));
          Mu = (double*)malloc(nxny * sizeof(double));
          RI = (double*)malloc(nxny * sizeof(double));
          O1 = (double*)malloc(nxny * sizeof(double));

          FILE  *f1;
          double d1;
          if (fopen_s(&f1, "ph-250x250.txt", "r") == 0)
          {
            for (i = 0; i < nxny; i++)
            {
              fscanf_s(f1, "%lf", &d1);
              Mu[i] = d1;
            }
            fclose(f1);
          }
          else
            MessageBox(NULL, (LPCWSTR)L"Problem at fopen_s 1!", (LPCWSTR)L"Error!", MB_OK);

          if (fopen_s(&f1, "ph-250x250-(2)-270.txt", "r") == 0)
          {
            for (i = 0; i < nxny; i++)
            {
              fscanf_s(f1, "%lf", &d1);
              RI[i] = (d1 < 0.0 ? 0.0 : d1);
            }
            fclose(f1);
          }
          else
            MessageBox(NULL, (LPCWSTR)L"Problem at fopen_s 2!", (LPCWSTR)L"Error!", MB_OK);


          for (i = 0; i < nxny; i++)
          {
            /*
            RI[i] = 0.0;
            */
            O1[i] = 0.0;
          }


          double *dc = NULL;
          dc = (double*)malloc(np * sizeof(double));

          dc[0] = -((((double)np) - 1.0) / 2.0)*((a + d) / d);
          for (i = 1; i < np; i++)
            dc[i] = dc[i - 1] + (a + d) / d;

          double x1;
          double x2;
          double y1;
          double y2;


          double *xl = NULL;
          double *yl = NULL;
          xl = (double*)malloc((ny + 1) * sizeof(double));
          yl = (double*)malloc((nx + 1) * sizeof(double));

          double xLoLimit;
          double xUpLimit;
          double yLoLimit;
          double yUpLimit;
          xLoLimit = -((((double)ny) - 1.0) / 2.0 + 0.5);
          xUpLimit =  ((((double)ny) - 1.0) / 2.0 + 0.5);
          yLoLimit = -((((double)nx) - 1.0) / 2.0 + 0.5);
          yUpLimit =  ((((double)nx) - 1.0) / 2.0 + 0.5);

          xl[0] = xLoLimit;
          for (i = 1; i < (ny + 1); i++)
            xl[i] = xl[i - 1] + 1.0;
          yl[0] = yLoLimit;
          for (i = 1; i < (nx + 1); i++)
            yl[i] = yl[i - 1] + 1.0;


          double xhrz;
          double yhrz;
          double xvrt;
          double yvrt;


          double *S1 = NULL;
          S1 = (double*)malloc(npnf * sizeof(double));
          for (i = 0; i < npnf; i++)
            S1[i] = 0.0;

          double f, m, b;
          double Sinf;
          double Sinfa;
          double Cosf;
          double Cosfa;


          double       xcm;
          double       ycm;
          unsigned int Lin;
          unsigned int Col;
          double       seg;

          double      *xV;
          double      *yV;
          xV = (double*)malloc((ny + 1) * (nx + 1) * sizeof(double));
          yV = (double*)malloc((ny + 1) * (nx + 1) * sizeof(double));
          vector<pair<double, double>> V;
          unsigned int nEl;

          unsigned int nyPlusOne  = ny + 1;
          unsigned int nxPlusOne  = nx + 1;
          unsigned int nyMinusOne = ny - 1;
          unsigned int nxMinusOne = nx - 1;
          double       Var1       = (((double)ny) - 1.0) / 2.0 + 0.5;
          double       Var2       = (((double)nx) - 1.0) / 2.0 + 0.5;

          unsigned int index1;
          unsigned int index2;
          unsigned int index3;
          unsigned int index4;
          double       maxS1 = -1.0;
          unsigned int imax;
          unsigned int jmax;

          double aux;
          for (j = 0; j < nf; j++)
          {
            f = j*c1;

            Sinf = sin(f); Sinfa = Sinf*a;
            Cosf = cos(f); Cosfa = Cosf*a;

            x1 = (-d)*(-Sinf);
            y1 = (-d)*( Cosf);

            for (i = 0; i < np; i++)
            {
              index1 = i*nf + j;

              Z1[index1].count     = 0;
              Z1[index1].ind       = NULL;
              Z1[index1].seg       = NULL;
              Z1[index1].SumOfSegs = 0.0;

              nEl = 0;

              x2 = Cosf*dc[i] - Sinfa;
              y2 = Sinf*dc[i] + Cosfa;

              if (x1 != x2)
              {
                if (y1 != y2)
                {
                  m = (y2 - y1) / (x2 - x1);
                  b = y1 - m*x1;

                  for (k = 0; k < nyPlusOne; k++)
                  {
                    xhrz = (xl[k] - b) / m;
                    yhrz =  xl[k];
                    if ((xhrz >= xLoLimit) && (xhrz <= xUpLimit) && (yhrz >= yLoLimit) &&
                                                                              (yhrz <= yUpLimit))
                    {
                      V.push_back(make_pair(xhrz, yhrz));
                      nEl++;
                    }
                  }
                  for (k = 0; k < nxPlusOne; k++)
                  {
                    xvrt = yl[k];
                    yvrt = m*yl[k] + b;
                    if ((xvrt >= xLoLimit) && (xvrt <= xUpLimit) && (yvrt >= yLoLimit) &&
                                                                              (yvrt <= yUpLimit))
                    {
                      V.push_back(make_pair(xvrt, yvrt));
                      nEl++;
                    }
                  }

                  sort(V.begin(), V.end());
                  for (k = 0; k < nEl; k++)
                  {
                    xV[k] = V[k].first;
                    yV[k] = V[k].second;
                  }

                  if (nEl >= 2)
                  {
                    Z1[index1].count = nEl - 1;
                    Z1[index1].ind = (unsigned int*)malloc(Z1[index1].count *
                                                                           sizeof(unsigned int));
                    Z1[index1].seg = (double*      )malloc(Z1[index1].count *
                                                                           sizeof(      double));

                    for (k = 1; k < nEl; k++)
                    {
                      xcm = (xV[k - 1] + xV[k]) / 2.0;
                      ycm = (yV[k - 1] + yV[k]) / 2.0;
                      Col = (int)floor(xcm + Var1);
                      if (Col > nyMinusOne)
                        Col = Col - 1;
                      Lin = (int)floor(Var2 - ycm);
                      if (Lin > nxMinusOne)
                        Lin = Lin - 1;
                      seg = sqrt(pow(xV[k] - xV[k - 1], 2) + pow(yV[k] - yV[k - 1], 2));
                      S1[index1] = S1[index1] + seg*Mu[Lin*ny + Col];

                      Z1[index1].ind[k - 1] = Lin*ny + Col;
                      Z1[index1].seg[k - 1] = seg;
                      Z1[index1].SumOfSegs += seg;
                    }
                    if (S1[index1] > maxS1)
                    {
                      maxS1 = S1[index1];
                      imax = i;
                      jmax = j;
                    }
                  }
                }
                else
                {
                  if ((y1 >= yLoLimit) && (y1 <= yUpLimit))
                  {
                    Z1[index1].count = ny;
                    Z1[index1].ind = (unsigned int*)malloc(Z1[index1].count *
                                                                           sizeof(unsigned int));
                    Z1[index1].seg = (double*      )malloc(Z1[index1].count *
                                                                           sizeof(      double));

                    ycm = y1;
                    Lin = (int)floor(Var2 - ycm);
                    if (Lin > nxMinusOne)
                      Lin = Lin - 1;
                    for (k = 1; k <= ny; k++)
                    {
                      xcm = (xl[k - 1] + xl[k]) / 2.0;
                      Col = (int)floor(xcm + Var1);
                      if (Col > nyMinusOne)
                        Col = Col - 1;
                      S1[index1] = S1[index1] + Mu[Lin*ny + Col];

                      Z1[index1].ind[k - 1] = Lin*ny + Col;
                      Z1[index1].seg[k - 1] = 1.0;
                      Z1[index1].SumOfSegs += 1.0;
                    }
                    if (S1[index1] > maxS1)
                    {
                      maxS1 = S1[index1];
                      imax = i;
                      jmax = j;
                    }
                  }
                }
              }
              else
              {
                if ((x1 >= xLoLimit) && (x1 <= xUpLimit))
                {
                  Z1[index1].count = nx;
                  Z1[index1].ind = (unsigned int*)malloc(Z1[index1].count *sizeof(unsigned int));
                  Z1[index1].seg = (double*      )malloc(Z1[index1].count * sizeof(     double));

                  xcm = x1;
                  Col = (int)floor(xcm + Var1);
                  if (Col > nyMinusOne)
                    Col = Col - 1;
                  for (k = 1; k <= nx; k++)
                  {
                    ycm = (yl[k - 1] + yl[k]) / 2.0;
                    Lin = (int)floor(Var2 - ycm);
                    if (Lin > nxMinusOne)
                      Lin = Lin - 1;
                    S1[index1] = S1[index1] + Mu[Lin*ny + Col];

                    Z1[index1].ind[k - 1] = Lin*ny + Col;
                    Z1[index1].seg[k - 1] = 1.0;
                    Z1[index1].SumOfSegs += 1.0;
                  }
                  if (S1[index1] > maxS1)
                  {
                    maxS1 = S1[index1];
                    imax = i;
                    jmax = j;
                  }
                }
              }

              V.clear();
            }
          }
			
          start2 = clock();


          unsigned int  ray1;
          unsigned int  ray2;

          double        Sit1;
          double        Sit2;
          double        Ratio1;
          double        coeff;
          double        coeff1;
          double        coeff2;
          unsigned int *pi1;
          unsigned int *pi2;
          double       *pd1;

          double        maxRay1;
          double        maxRay2;
          double        maxRay3;
          double        sumExt;
          double        sumOFS;

          unsigned int  I1;

          int           z;
          int           T;



          if (fopen_s(&f1, "pairs-270x270.txt", "r") != 0)
            MessageBox(NULL, (LPCWSTR)L"Problem at fopen_s!", (LPCWSTR)L"Error!", MB_OK);
			
          unsigned int *vRay1 = (unsigned int*)malloc(6282674 * sizeof(unsigned int));
          unsigned int *vRay2 = (unsigned int*)malloc(6282674 * sizeof(unsigned int));

          for (z = 0; z < 6282674; z++)
          {
            fscanf_s(f1, "%d", &ray1);
            fscanf_s(f1, "%d", &ray2);
            vRay1[z] = ray1;
            vRay2[z] = ray2;
          }

          fclose(f1);
			
          start2 = clock();

          int *Mzeros = (int*)malloc(nxny * sizeof(int));
          for (i = 0; i < nxny; i++)
            Mzeros[i] = 1;
          for (i = 0; i < npnf; i++)
          {
            if (S1[i] == 0.0)
              for (k = 0; k < Z1[i].count; k++)
                Mzeros[Z1[i].ind[k]] = 0;
          }

          for (i = 0; i < npnf; i++)
          {
            Z1[i].Ind = NULL;
            Z1[i].Seg = NULL;
            Z1[i].Count = 0;
            if (Z1[i].count > 0)
            {
              Z1[i].Ind = (unsigned int*)malloc(Z1[i].count * sizeof(unsigned int));
              Z1[i].Seg = (double*      )malloc(Z1[i].count * sizeof(      double));
              for (k = 0; k < Z1[i].count; k++)
                if (Mzeros[Z1[i].ind[k]] == 1)
                {
                  Z1[i].Ind[Z1[i].Count] = Z1[i].ind[k];
                  Z1[i].Seg[Z1[i].Count] = Z1[i].seg[k];
                  Z1[i].Count++;
                }
            }
          }

			

          i  = 0;
          I1 = 125000;
          for (z = I1; z != 0; z--)
          {
            ray1 = vRay1[i];
            ray2 = vRay2[i];
            i++;

            if ((S1[ray1] > 0.0) && (S1[ray2] > 0.0))
            {
              index1 = Z1[ray1].Count;
              index2 = Z1[ray2].Count;

              Sit1 = 0.0;
              pi1 = Z1[ray1].Ind;
              pd1 = Z1[ray1].Seg;
              for (k = 0; k < index1; k++)
              {
                Sit1 += (*pd1) * RI[*pi1];
                pi1++;
                pd1++;
              }

              Sit2 = 0.0;
              pi1 = Z1[ray2].Ind;
              pd1 = Z1[ray2].Seg;
              for (k = 0; k < index2; k++)
              {
                Sit2 += (*pd1) * RI[*pi1];
                pi1++;
                pd1++;
              }

              Ratio1 = S1[ray1] / S1[ray2];
              coeff = (Ratio1 * Sit2 - Sit1) / (1.0 + Ratio1);
              coeff1 = coeff / Sit1;
              coeff2 = coeff / Sit2;

              pi1 = Z1[ray1].Ind;
              for (k = 0; k < index1; k++)
              {
                RI[*pi1] += (1.0)*coeff1*RI[*pi1];
                pi1++;
              }


              pi1 = Z1[ray2].Ind;
              for (k = 0; k < index2; k++)
              {
                RI[*pi1] -= (1.0)*coeff2*RI[*pi1];
                pi1++;
              }
            }
            else
            {
              z++;
            }
          }

          finish = clock();

          duration1 = (double)(start2 - start1) / CLOCKS_PER_SEC;
          duration2 = (double)(finish - start2) / CLOCKS_PER_SEC;



          for (i = 0; i < npnf; i++)
          {
            free(Z1[i].ind);
            free(Z1[i].seg);
            free(Z1[i].Ind);
            free(Z1[i].Seg);
          }
          free(Z1);
          free(Mu);
          free(O1);
          free(dc);
          free(xl);
          free(yl);
          free(S1);
          free(xV);
          free(yV);
          free(vRay1);
          free(vRay2);
          free(Mzeros);


          min1 = RI[0];
          for (i = 0; i < nxny; i++)
            if (RI[i] < min1)
              min1 = RI[i];

          max1 = RI[0];
          for (i = 0; i < nxny; i++)
            if (RI[i] > max1)
              max1 = RI[i];


          char *buffer1 = NULL;

          buffer1 = (char*)malloc(_CVTBUFSIZE);

          if (fopen_s(&f1, "fout1.txt", "w+t") == 0)
          {
            for (i = 0; i < nxny; i++)
            {
              sprintf_s(buffer1, _CVTBUFSIZE, "%f\n", RI[i]);
              fputs(buffer1, f1);
            }
            fclose(f1);
          }
          else
            MessageBox(NULL, (LPCWSTR)L"Problem at fopen_s 3!", (LPCWSTR)L"Error!", MB_OK);

          free(buffer1);



          if (pixels)
            free(pixels);
          pixels = (unsigned char*)malloc(3 * nx * ny + 2 * nx);

          unsigned char c;
          for (i = 0; i < nx; i++)
          {
            for (j = 0; j < ny; j++)
            {
              c = (unsigned char)((RI[i*ny + j] / max1)*255.0);
              pixels[i * 3 * ny + i * 2 + 3 * j    ] = c;
              pixels[i * 3 * ny + i * 2 + 3 * j + 1] = c;
              pixels[i * 3 * ny + i * 2 + 3 * j + 2] = c;
            }
          }



          SetCursor(LoadCursor(NULL, IDC_WAIT));
          InvalidateRect(hWnd, NULL, TRUE);
          Status = 1;
          MSG msg1;
          msg1.hwnd = hWnd;
          msg1.message = WM_PAINT;
          DispatchMessage(&msg1);
        }
        break;
        case IDM_ABOUT:
          DialogBox(hInst, MAKEINTRESOURCE(IDD_ABOUTBOX), hWnd, About);
        break;
        case IDM_EXIT:
          DestroyWindow(hWnd);
        break;
        default:
          return DefWindowProc(hWnd, message, wParam, lParam);
      }
    }
    break;
    case WM_PAINT:
    {
      PAINTSTRUCT ps;
      HDC hdc = BeginPaint(hWnd, &ps);

      if (Status == 1)
      {
        bmih.biSize = sizeof(BITMAPINFOHEADER);
        bmih.biWidth = (int)ny;
        bmih.biHeight = -(int)nx;
        bmih.biPlanes = 1;
        bmih.biBitCount = 24;
        bmih.biCompression = BI_RGB;
        bmih.biSizeImage = 0;
        bmih.biXPelsPerMeter = 10;
        bmih.biYPelsPerMeter = 10;
        bmih.biClrUsed = 0;
        bmih.biClrImportant = 0;


        ZeroMemory(&dbmi, sizeof(dbmi));
        dbmi.bmiHeader = bmih;
        dbmi.bmiColors->rgbBlue = 0;
        dbmi.bmiColors->rgbGreen = 0;
        dbmi.bmiColors->rgbRed = 0;
        dbmi.bmiColors->rgbReserved = 0;
        bits = (void*)&(pixels[0]);


        hbmp = CreateDIBSection(hdc, &dbmi, DIB_RGB_COLORS, &bits, NULL, 0);

        if (hbmp == NULL)
          MessageBox(hWnd, (LPCWSTR)L"Couldn't create bitmap!", (LPCWSTR)L"Error!", MB_OK);
        memcpy(bits, pixels, 3 * nx * ny + 2 * nx);


        hdcMem = CreateCompatibleDC(hdc);
        hbmOld = SelectObject(hdcMem, hbmp);
        GetObject(hbmp, sizeof(bmp), &bmp);
        BitBlt(hdc, 0, 0, bmp.bmWidth, bmp.bmHeight, hdcMem, 0, 0, SRCCOPY);
        SelectObject(hdcMem, hbmOld);
        DeleteDC(hdcMem);

        RECT r1, r2, r3, r4;
        r1.left   = 0;
        r1.top    = nx + 10;
        r1.right  = ny;
        r1.bottom = nx + 40;
        r2.left   = 0;
        r2.top    = nx + 40;
        r2.right  = ny;
        r2.bottom = nx + 70;
        r3.left   = 0;
        r3.top    = nx + 70;
        r3.right  = ny;
        r3.bottom = nx + 100;
        r4.left   = 0;
        r4.top    = nx + 100;
        r4.right  = ny;
        r4.bottom = nx + 130;

        WCHAR buffer1[50];
        int len;
        len = swprintf(buffer1, 50, L"Time 1: %12.2f ", duration1);
        DrawText(hdc, (LPTSTR)buffer1, len, &r1, DT_RIGHT);

        len = swprintf(buffer1, 50, L"Time 2: %12.2f ", duration2);
        DrawText(hdc, (LPTSTR)buffer1, len, &r2, DT_RIGHT);

        len = swprintf(buffer1, 50, L"   min: %12.2f ", min1     );
        DrawText(hdc, (LPTSTR)buffer1, len, &r3, DT_RIGHT);

        len = swprintf(buffer1, 50, L"   max: %12.2f ", max1     );
        DrawText(hdc, (LPTSTR)buffer1, len, &r4, DT_RIGHT);
      }


      EndPaint(hWnd, &ps);
    }
    break;
    case WM_DESTROY:
      if (hbmp)
        DeleteObject(hbmp);
      PostQuitMessage(0);
    break;
    default:
      return DefWindowProc(hWnd, message, wParam, lParam);
  }
  return 0;
}

// Message handler for about box.
INT_PTR CALLBACK About(HWND hDlg, UINT message, WPARAM wParam, LPARAM lParam)
{
  UNREFERENCED_PARAMETER(lParam);
  switch (message)
  {
    case WM_INITDIALOG:
      return (INT_PTR)TRUE;

    case WM_COMMAND:
      if (LOWORD(wParam) == IDOK || LOWORD(wParam) == IDCANCEL)
      {
        EndDialog(hDlg, LOWORD(wParam));
        return (INT_PTR)TRUE;
      }
    break;
  }
  return (INT_PTR)FALSE;
}
\end{lstlisting}
\subsection{Generation of Non-overlapping X-rays}
When describing the second randomized algorithm, we have explained that the two X-rays chosen at each iteration have to be such that they don't have share any element of the reconstruction matrix. For the example of the 250 by 250 phantom that we have tested, we have generated enough pairs such that up to 1000000 iterations can be run on the shown example. Precisely, 6282674 pairs of X-rays have been generated, for the special geometry and parameter values of the shown example with 270 projections views. We have calculated that exactly these number of pairs are needed in order to run up to 1000000 iterations for the shown example with 270 projection views. That is, 5282674 pairs from this pool of pairs will not satisfy the condition
\begin{displaymath}
{\it if }((S1[ray1] > 0.0){\it  \&\& }(S1[ray2] > 0.0)).
\end{displaymath}
Before the correction algorithm starts, as one can see, the pairs are loaded in memory and used during the correction algorithm.
\section{Conclusions}
We have proposed fast corection of analytical algorithms for sparse view X-ray CT, that provide significantly better reconstructions as compared to the results obtained by the analytical algorithms alone, and thus can be used for speeding-up the experiment, as less number of views are required in order to obtain comparable quality of reconstruction. One important thing that needs to be done with the proposed randomized iterative algorithms is to study their convergence and define precise stopping criteria to be used in practice.

\end{document}